\documentclass[10pt,twocolumn,letterpaper]{article}

\usepackage{iccv}
\usepackage{times}
\usepackage{epsfig}
\usepackage{graphicx}
\usepackage{amsmath}
\usepackage{amssymb}
\usepackage{verbatim}
\usepackage{colortbl}
\usepackage{tabu}
\usepackage{booktabs}

\usepackage{color}
\usepackage{multirow}

\newcommand{\rev}[1]{{\color{black}{#1}}}

\newcommand\blfootnote[1]{%
  \begingroup
  \renewcommand\thefootnote{}\footnote{#1}%
  \addtocounter{footnote}{-1}%
  \endgroup
}

\usepackage[pagebackref=true,breaklinks=true,letterpaper=true,colorlinks,bookmarks=false]{hyperref}

\iccvfinalcopy % *** Uncomment this line for the final submission

\ificcvfinal\fi
\begin{document}

%%%%%%%%% TITLE
\title{Understanding Infographics through Textual and Visual Tag Prediction}
%\title{Learning Visual Hashtags for Infographics}

\author{Zoya Bylinskii\textsuperscript{1}*	\quad Sami Alsheikh\textsuperscript{1}* \quad Spandan Madan\textsuperscript{2}* \quad Adri\`{a} Recasens\textsuperscript{1}*\\
Kimberli Zhong\textsuperscript{1} \quad Hanspeter Pfister\textsuperscript{2} \quad Fredo Durand\textsuperscript{1} \quad Aude Oliva\textsuperscript{1} \\
\textsuperscript{1} Massachusetts Institute of Technology \quad
\textsuperscript{2} Harvard University\\
{\tt\small \{zoya,alsheikh,recasens,kimberli,fredo,oliva\}@mit.edu} \\
{\tt\small \{spandan\_madan,pfister\}@seas.harvard.edu}
% For a paper whose authors are all at the same institution,
% omit the following lines up until the closing ``}''.
% Additional authors and addresses can be added with ``\and'',
% just like the second author.
% To save space, use either the email address or home page, not both
%\and
%Second Author\\
%Institution2\\
%First line of institution2 address\\
%{\tt\small secondauthor@i2.org}
}

\maketitle
%\thispagestyle{empty}

%%%%%%%%% ABSTRACT
\begin{abstract}
We introduce the problem of visual hashtag discovery for infographics:  extracting visual elements from \rev{an} infographic that are diagnostic of its topic. Given an infographic as input, our computational approach automatically outputs textual and visual elements predicted \rev{to be} representative of the infographic content. Concretely, from a curated dataset of 29K large infographic images sampled across 26 categories and 391 tags, we present an automated two step approach. First, we extract the text from an infographic and use it to predict text tags indicative of the infographic content. And second, we use these predicted text tags \rev{as a supervisory signal} to localize the most diagnostic visual elements from within the infographic i.e. visual hashtags. We report performances on a categorization and multi-label tag prediction problem and compare our proposed visual hashtags to human annotations.
\end{abstract}
% visual hashtags differ from summaries or thumbnails as they are not a smaller or abstract version of a large image, but
% We report performances on a categorization and multi-label tag prediction problem, and compare results of our autmoatically extracted visual hashtags to human annotations.
%%%%%%%%% BODY TEXT
\section{Introduction}

\emph{If a hashtag can be worth 140 characters, how much is a visual hashtag worth?} While text can be used to clearly convey a short message, a meaningful icon conveys the gist of a webpage or poster right away, grabbing attention while helping store the message in memory \cite{borkin2016beyond}. Identifying these visual regions requires an understanding of both the textual and visual content of the infographic.  In this paper, we introduce a system that identifies these ``visual hashtags", iconic image regions    that represent key topics of an infographic. \rev{For instance, given an infographic with topics ``economy'' and ``environment", relevant visual hashtags could be crops showing a coin (for economy) or the earth (for environment).\blfootnote{* Indicates equal contribution.}} 

%\emph{If a hashtag can be worth 140 characters, how much is a visual hashtag worth?} While text can be used to clearly convey a message, visual elements are what grab an observer's attention and stay in memory as visual hooks or associations \cite{borkin2016beyond}. It is the visual elements on a webpage or poster that draw us in and can, in an instant, give us a gist of what the content is about. Effective visual elements can guide an observer's understanding.

\begin{figure}[h]
\caption{Our proposed approach for generating visual hashtags: the text in an infographic predicts the tags, the visual model fires on the patches that most activate for this tag, and a segmentation pipeline is run to extract the representative visual elements from the highly activating regions of the infographic. The result is automatically generated by our model.} % (infographic cropped to fit).
\centering
\includegraphics[height=10cm]{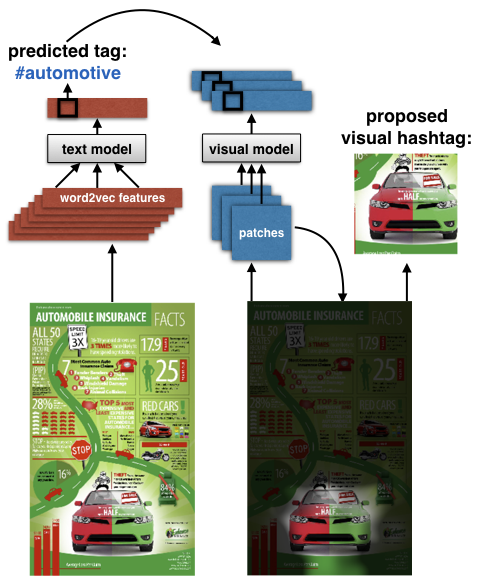}
\label{fig:pipeline}
\end{figure}

Infographics are visual encodings of visual and textual media, including graphs, visualizations, and graphic designs. They are specifically designed to provide an effective visual digest with the intent of delivering a message. Tags can serve as key words describing this message to facilitate data organization, retrieval from large databases, and sharing on social media.
%  (Fig.~\ref{fig:visually})

%They are specifically designed to provide an effective visual digest with the intent of delivering a message, communicating a concept, or telling a story.  Tags can thereby serve as concise summaries of the topics in infographics to facilitate data organization, retrieval from large databases, and sharing across social communities. 

Analogously, we propose an \textbf{effective visual digest} of infographics via \textbf{visual hashtags}. Instead of providing visual summaries or thumbnails of the whole infographic, visual hashtags correspond to specific visual concepts or topics inside the infographic's rich visual space. We introduce a computational system that, given an infographic as input, produces discriminative textual and visual hashtags. \rev{Just as YouTube videos use representative frames as thumbnails, we identify relevant crops of an infographic as a ``preview" of its content. Such thumbnails may aid in retrieval applications (e.g. organizing and visualizing large infographic collections from a webpage or file system).}
We evaluate the quality of visual hashtags by comparing the system's output to the image regions humans \rev{box as relevant} to a particular textual tag on a given image.

%Analogously, we propose an \textbf{effective visual digest} of infographics through \textbf{visual hashtags}. Instead of providing visual summaries or thumbnails of the whole infographic, we intend our visual hashtags to narrow in on the key visual concepts or topics inside the rich visual space of an infographic. Given an infographic as input, we introduce a computational system to output the textual hashtags and visual hashtags predicted most representative of the infographic. 

Unlike most natural images, infographics often contain embedded text that provides meaningful context for the visual content. We leverage this text to first make category (topic) and tag (sub-topic) predictions. We then use these predictions to constrain and disambiguate the automatically extracted visual features.

\begin{figure*}
\caption{Our visual network learns to associate visual elements like pictographs with category labels. We show the activations of our visual network conditioned on different category labels for the same infographic. Allowing the text in an infographic to make the high-level category predictions constrains the visual features to focus on the relevant image regions, in this case \emph{Environment}, the correct category for the image. Image source: \url{http://oceanservice.noaa.gov/ocean/earthday-infographic-large.jpg}}
\centering
\includegraphics[width=1\linewidth]{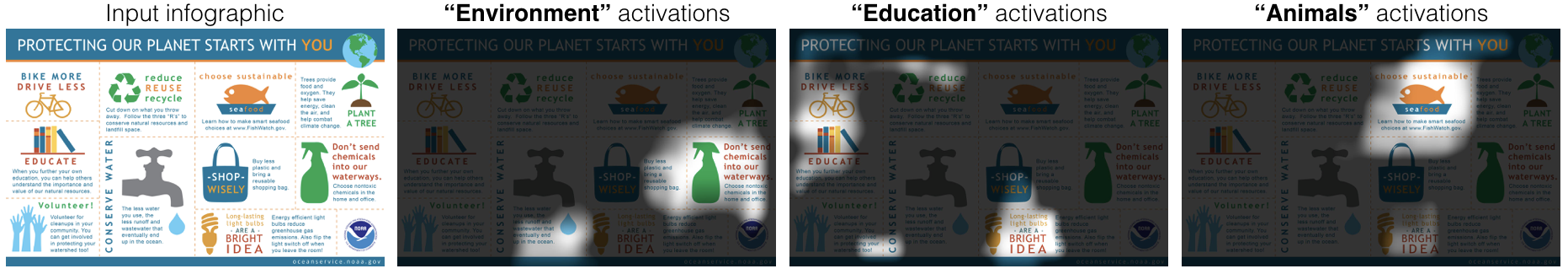}
\label{fig:earthday}
%\vspace{-0.9cm}
\end{figure*}

%This disambiguation is a key step in identifying the most diagnostic regions of an infographic. For instance, if the text on an infographic predicts the category \emph{Environment}, then the system can condition visual object proposals on the presence of this topic and highlight regions relating to \emph{Environment}. In the case of the infographic in Fig. ~\ref{fig:earthday}, this disambiguation allows our system to focus on the water droplet and spray bottle, as opposed to the books and light bulb highlighted by \emph{Education}.\\
\rev{
This disambiguation is a key step in identifying the most diagnostic regions of an infographic. For instance, in Fig. ~\ref{fig:earthday} which contains diverse visual elements, if a predicted text tag is \emph{Environment} then the system can condition visual object proposals on this topic and focus on related regions like the water droplet and spray bottle. On the other hand, if the predicted text tag is \emph{Education}, the system can condition proposals on regions like the book. Thus, we can use the predicted text tags as a kind of supervisory signal for the visual model, to identify visual regions indicative of the different topics in the infographic.}

% for example.

%For instance, if the text on an infographic predicts the tag \emph{\#social media}, then the visual object proposals can be conditioned on the presence of this tag and produce visual \emph{\#social media} hashtags (Fig.~\ref{fig:pipeline}). This way, our system returns the most relevant elements

% Unlike most natural images, infographics often contain embedded text that provides meaningful description and context for the visual content. Here, we leverage this text to make category and tag predictions and to constrain and disambiguate the visual features automatically extracted. For instance, if the text on an infographic predicts the topic \emph{\#politics}, then the visual object proposals can be conditioned on the presence of this tag to look for the most representative visual features and produce visual \emph{\#politics} hashtags.

\begin{comment}
\begin{figure}
\caption{}
\centering
\includegraphics[width=1\linewidth]{earthday.png}
\label{fig:earthday}
\end{figure}
\end{comment}

\begin{comment}
\begin{figure}[h]
\caption{From an input infographic image, our model automatically predicts text tags and extract visual hashtags.}
\centering
\includegraphics[width=0.5\textwidth]{hashtags.png}
\label{fig:pipeline}
\end{figure}
\end{comment}

\textbf{Approach:} We present our tagging application on a dataset of 29K infographics scraped from \emph{Visually} (\url{http://visual.ly/view}). Each infographic comes with a designer-assigned category label, multiple tags, and other meta-data (Sec.~\ref{sec:dataset}). We achieve prediction accuracy of \rev{46\%} when predicting the top category out of 26 \rev{categories}. For text tags, we achieve \rev{48.2\%} top-1 average precision at predicting at least one of the possible few tags for an image out of 391 possible tags. These predictions are driven by text that we automatically extracted from the infographics and post-processed with a single-hidden-layer neural network (Sec.~\ref{sec:text2labels}). Separately, we train category and tag prediction from image patches using a deep multiple-instance learning framework (Sec.~\ref{sec:patches2labels}). At test time, we run our patch-based visual network densely over an infographic, constrained to the tags predicted by the text network, to generate visual region proposals associated to the text tags. These proposals are then fed to a deep mask segmentation pipeline to generate the final visual hashtags (Fig.~\ref{fig:pipeline}).

% NOTE - Sami: I am worried saying that we "introduce the problem of visual hashtag discovery" they will then criticize our problem statement / metrics / evaluation a lot I think. So maybe we can re-word this or be more confident in the problem statement.
\textbf{Contributions:} We introduce the problem of visual hashtag discovery, which consists of extracting diagnostic visual regions for particular topics. 
%\textcolor{red}{Visual hashtags are not object detections ; they are representative elements, icons, and pictographs that are visually meaningful in infographics. Aude: this sentence requires a better explanation, too vague. or remove the sentence, it is not needed here}
We demonstrate the utility of a patch-based, deep multiple instance approach for the processing of intractably large (up to \rev{8000} pixels/side) and visually rich images. Unlike approaches that use text outside of an image for visual recognition tasks, we show the power of extracting text from \emph{within} the image itself for facilitating visual recognition. On a novel curated dataset of 29K infographics, we report performances on a categorization problem and a multi-label tag prediction problem, and show results of our automatically extracted visual hashtags. %\textcolor{red}{here you need to be a little more precise about the quality of results, "show results" is too vague.}

%\newpage
\section{Related work}

%\begin{itemize}
%\item Cognitive/perception work on effectiveness of visual elements
%\item Visualizations and infographics in general and work on them
%Many works have shown that observers attention can be grabbed in a consistent manner based on features like saliency, object importance or memorability. In the domain of infographic for instance, \cite{borkin2013makes} found that observers are highly consistent in which visualizations they find memorable and forgettable, and importantly that recognizable object enhance memorability of the whole infographic \cite{isola2014makes}. 

% POSSIBILE WAY TO FRAME - Identifying the region of interest in an image is an active area of research in computer vision. Research has shown that an observer's attention can be captured in a consistent manner based on features like saliency, memorability and object importance.

%\item Visual summarization, thumbnailing and how our problem is different 
 
%\item vision for digitally born media
\rev{Conventionally, computer vision research has focused mostly on understanding natural images and scenes, while very little work has been done on digitally born media. Some work has been present in \cite{seo2014diagram} where the authors use computer vision techniques for geometry diagrams, and more recently in \cite{kembhavi2016eccv} where the authors use graph structures to syntactically parse diagrams. In a similar vein, \cite{zitnick2016adopting} show that simpler, abstract digital images can be used in place of natural images to understand the semantic relationship between visual media and their natural language representation. However, to the best of our knowledge there is no work on automated understanding of infographics using computer vision techniques.\\
%\item reviewers papers etc
\indent Our task of text tag prediction for images is similar to that presented in \cite{Denton:2015:UCH:2783258.2788576}, however we attempt it on infographic images as opposed to natural images. Also, unlike \cite{Denton:2015:UCH:2783258.2788576}, where the authors trained a joint embedding of visual and text features, we solve the problem using just the visual features of an image. To work around the large size of the infographic images, we use a variant of multiple-instance learning approach \cite{dietterich1997solving}.\\
\indent We also predict text tags using the text extracted from within these images, which has not been tried before to the best of our knowledge. To obtain a distributed representation for the extracted text, we used the mean word2vec~\cite{DBLP:journals/corr/abs-1301-3781} representation, as suggested by \cite{wu2015deep}. We also tried other representations like the glove embedding \cite{pennington2014glove}, and tweet2vec \cite{dhingra-EtAl:2016:P16-2}.

\indent In this paper, we present a method to extract visual hashtags from infographics using only image-level tags. This weakly suppervised learning scheme is similar to \cite{zhou2016learning}, where the category labels are used to estimate the location of the elements in the image. However, unlike \cite{zhou2016learning}, we combine this weakly supervised model with a tag classifier based in the extracted text to improve the final prediction. 

}
%\item Multiple-instance learning
%\end{itemize}
%The Multiple Instance Learning approach \cite{dietterich1997solving} \cite{andrews2003support} serves as a great way to approach weakly supervised problems in machine learning. It has been used successfully in many domains across vision including: tracking \cite{babenko2009visual} \cite{babenko2011robust}, action recognition \cite{ali2010human} and category prediction from keywords \cite{vijayanarasimhan2008keywords}. Recently, \cite{wu2015deep} showed that the framework can be used in conjunction with a deep neural network to predict categories from noisy metadata scraped from the internet.

\section{Infographics dataset} \label{sec:dataset}
%\subsection{Dataset - Comparison to other datasets}
We scraped 63,885 static infographic images from the \emph{Visually} website, a community platform for hand curated visual content.
Each infographic is hand categorized, tagged, and described by the designer, making it a rich source of annotated images. Despite the difference in visual content, compared to other scene text datasets such as ICDAR 03 \cite{lucas2003icdar}, ICDAR 15 \cite{karatzas2015icdar}, COCO-Text \cite{veit2016coco} and VGG SynthText in the wild \cite{Gupta16}, the \emph{Visually} dataset is similar in size and richness of text annotations, with metadata including labels for 26 categories (available for 90.21\% of the images), 19K tags (for 76.81\% of the images), titles (99.98\%) and descriptions (93.82\%). %Viewer likes, comments, and shares were also collected. %For a subset of 1193 images, full transcripts are available.

\begin{comment}
\begin{figure*}
\caption{(a) Sample data available from \emph{Visually}. We scraped over 63K infographics containing category, tags, and other annotations. (b) A few infographics from the dataset demonstrating the mix of textual and visual regions, the richness of visual content, and styles.}
\centering
(a)\includegraphics[height=5cm]{infovis1.png}
(b)\includegraphics[height=5cm]{infovis2.png}
\label{fig:visually}
\end{figure*}
\end{comment}

We curated a subset of this 63K dataset to obtain a representative subset of 28,973 images (Table~\ref{tab:datasetstats}). Uploaded tags are free text, so many of the original tags are either semantically redundant or have too few instances. Redundant tags were merged using WordNet~\cite{miller1995wordnet} and manually, and only the 391 tags with at least 50 image instances each were retained. To produce the final 29K dataset, we further filtered images to contain a category annotation, at least one of the 391 tags.  99.6\% of these images had visual aspect ratio between 1:5 and 5:1. Of this dataset, 10\% was held out as our test set, and the remaining 26K images were used for training our text and visual models. \rev{For 330 of the test images, we collected additional crowdsourced annotations in order to have ground truth visual hashtags for evaluation}.
% To evaluate our visual hashtags, for which no ground truth data is available. We ran our crowdsourcing studies on a subset of 330 images out of our test set, manually curated to suit the size and format constraints of our user interfaces. We also made sure transcripts were available for these images for our analyses.}

%For 330 of the test images, we collected additional crowdsourced textual tags and visual element bounding boxes for finer-grained evaluation (Sec.~\ref{sec:userstudies}).

% \begin{table}[h]
% \small
% \centering
% \caption{\emph{Visually} dataset statistics. We curated the original 63K infographics available on \emph{Visually} to produce a representative dataset with consistent tags and sufficient instances per tag.} \label{tab:datasetstats}
% \begin{tabular}{| p{0.8cm} | p{0.85cm} | p{1.4cm} | p{0.8cm} | p{1.2cm} | p{1.1cm} |}
% \hline
%   Dataset & \#of categ. & images per category & \#tags & images per tag & tags per image \\ \hline
%   63K (full) & 26 & min=184 max=9481 mean=2235 & 19,469 & min=1 max=3784 mean=7.8 & min=0
% max=10
% mean=3.7 \\ \hline
%   29K (clean) & 26 & min=118 max=4469 mean=1114 & 391 & min=50 max=2331 mean=151 & min=1 max=9 mean=2.1 \\ \hline
% \end{tabular} 
% \end{table}

\begin{table}[]
\small
\centering
\caption{\emph{Visually} dataset statistics. We curated the original 63K infographics available on \emph{Visually} to produce a representative dataset with consistent tags and sufficient instances per tag.} \label{tab:datasetstats}
\hspace*{-0.3cm}
\begin{tabular}{c c c c c c}
\hline
Dataset                                               & \begin{tabular}[c]{@{}c@{}}\# of\\ categ.\end{tabular} & \begin{tabular}[c]{@{}c@{}}Images per\\ categ.\end{tabular}          & \begin{tabular}[c]{@{}c@{}}\# of\\ tags\end{tabular} & \begin{tabular}[c]{@{}c@{}}Images\\ per tag\end{tabular}              & \begin{tabular}[c]{@{}c@{}}Tags per\\ Image\end{tabular}          \\ \hline
\begin{tabular}[c]{@{}c@{}}63k\\ (full)\end{tabular}  & 26                                                     & \begin{tabular}[c]{@{}c@{}}min=184\\ max=9481\\ mean=2235\end{tabular} & 19,469                                               & \begin{tabular}[c]{@{}c@{}}min=1\\ max=3784\\ mean=7.8\end{tabular}  & \begin{tabular}[c]{@{}c@{}}min=0\\ max=10\\ mean=3.7\end{tabular} \\ \hline
\begin{tabular}[c]{@{}c@{}}29k\\ (clean)\end{tabular} & 26                                                     & \begin{tabular}[c]{@{}c@{}}min=118\\ max=4469\\ mean=1114\end{tabular} & 391                                                  & \begin{tabular}[c]{@{}c@{}}min=50\\ max=2331\\ mean=151\end{tabular} & \begin{tabular}[c]{@{}c@{}}min=1\\ max=9\\ mean=2.1\end{tabular}  \\ \hline
\end{tabular}
%\vspace{-0.2cm}
\end{table}

\begin{comment}
\begin{table}[]
\small
\centering
\caption{\emph{Visually} dataset statistics. We curated the original 63K infographics available on \emph{Visually} to produce a representative dataset with consistent tags and sufficient instances per tag.} \label{tab:datasetstats}
\hspace*{-0.3cm}
\begin{tabu} to \linewidth {c c c c c c}
\toprule
Dataset & \# of categ. & Images per categ. & \# of tags & Images per tag & Tags per image   
\bottomrule
\end{tabu}
%\vspace{-0.2cm}
\end{table}
\end{comment}

%\subsection{Dataset statistics 63k}
%The number of images in each category range from 184 to 9481 images, with a mean of 2235.69. There are 19469 unique tags, with the number of image with each tag ranging from 0 to 3784, and an average of 7.8. Number of tags per image range from 0 to 10, and have an average of 3.75 tags/image.

%\subsection{Dataset statistics 30k}
%The number of images in each category range from 118 to 4469 images, with a mean of 1114.5. There are 391 unique tags, with the number of image with each tag ranging from 50 to 2331, and an average of 151.03. Number of tags per image range from 1 to 9, and have an average of 2.1 tags/image.
\section{Approach}

Given an infographic as input, our goal is to predict one or more text tags and \rev{visual hashtags} that are diagnostic of the \rev{topics depicted in the infographic}. We split this problem into two steps: (1) predicting the text tags for an infographic, and (2) using the predicted text tags to localize the most representative visual regions. 

Infographics are composed of a mix of text and visual elements, which combine to generate the message of the infographic. Given that the text is a very strong cue for the topic, we use it to provide context - a sort of supervisory signal - for the visual hashtag predictions. We use the text features to infer the category and tags for the infographic, and given these labels, we ask the visual model to predict the most confident visual regions \rev{indicative of these labels}. Learning a mapping directly from visual features to labels is a more ambiguous problem: not all topics are represented visually, and not all visual elements are relevant to the topic of the infographic (Sec.~\ref{sec:tagprediction}). Textual features help to disambiguate the mapping between visual regions and topics. \rev{Importantly, the text we use for prediction is extracted from within the image using optical character recognition.}
%Pictographs, photographic elements, or other imagery may be harder to directly map onto a topic or label, without the context of the text. Therefore, we harness our text predictions  

\begin{figure}
\caption{Our proposed \rev{training} approach separately samples and processes visual and text regions from an infographic to predict labels automatically. Bags of patches are sampled in a multiple instance learning formulation, and their predictions are averaged to produce the final classification. Text regions are automatically localized, extracted, and converted into word2vec representations. The average word2vec representation is then fed into a single hidden layer neural network to produce the final classification. }
\centering
\includegraphics[width=0.85\linewidth]{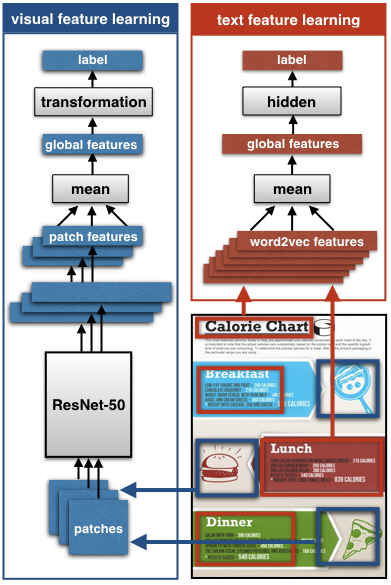}
\label{fig:architecture}
\end{figure}

\begin{figure*}
\caption{Samples of visual hashtags extracted for different concepts.}
\centering
\includegraphics[width=1\linewidth]{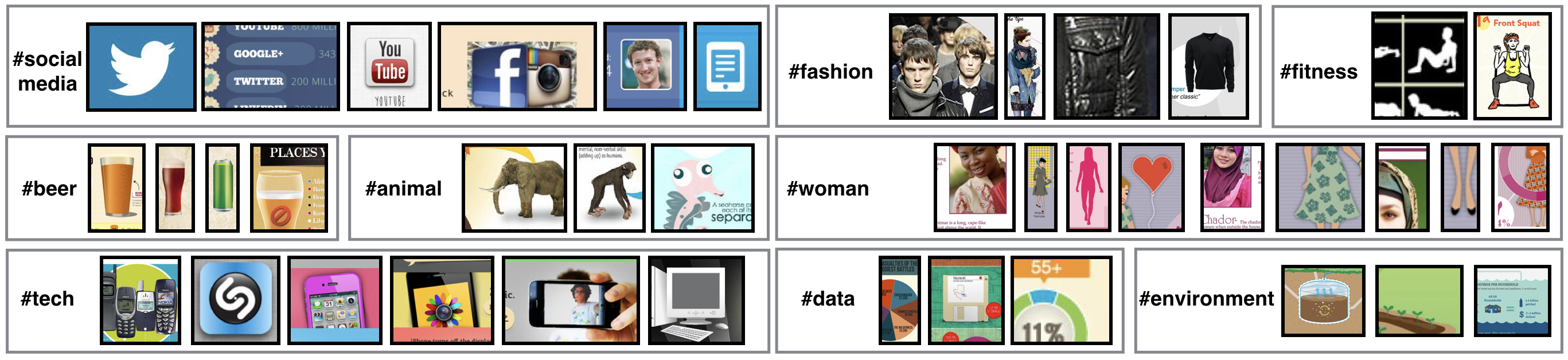}
\label{fig:vishashtags}
\end{figure*}

\subsection{Text to labels}\label{sec:text2labels}

Given an infographic encoded as a bitmap as input, we detected and extracted (i.e., optical character recognition) the text, and then used the text to predict labels for the whole infographic. These labels come in two forms: either a single category per infographic (1 of 26), or multiple tags per infographic (out of a possible 391 tags).  

\textbf{Automatic text extraction:} We used the stand-alone text spotting system of Gupta et al.~\cite{Gupta16} to discover text regions in our infographics. We automatically cleaned the text using spell checking and dictionary constraints in addition to the ones already in~\cite{Gupta16} to further improve results. On average, we extracted 95 words per infographic (capturing the title, paragraphs, annotations, and other text).

\textbf{Feature learning with text:} For each extracted word, we computed a 300-dimensional word2vec representation \cite{DBLP:journals/corr/abs-1301-3781}. The mean word2vec of the bag of extracted words was used as the distributed representation for the extracted text of the whole image (the global feature vector of the text). 
%A simple one hidden layer neural network was used for predicting the category and tag labels for each infographic (Fig.~\ref{}). 
%We found the mean word2vec representation of the bag of extracted words to work best as the input to the neural network.
% Inspired by \cite{wu2015deep}, a While \cite{wu2015deep} show great results with the Multiple Instance Learning approach for automatic keyword annotation for images, we failed to obtain good results with this approach.
%Instead, for every infographic, the mean of the word2vec representations of the bag of extracted words from it was used as its descriptor. 
This mean word2vec representation was fed into two single-hidden-layer neural networks for predicting the category and tags of each infographic. 
Category prediction was set up as a multi-class problem, where each infographic belongs to 1 of 26 categories. Tag prediction was set up as a multi-label problem with 391 tags, where each infographic could have multiple tags (Table~\ref{tab:datasetstats}). 
The network architecture is the same for both tasks and is depicted in the red box in Fig.~\ref{fig:architecture}, where the label is either one category or multiple tags. We used 26K labeled infographics for training and the rest for testing.
%The deep network contains one input layer, one hidden layer with ReLU activation function, and one output layer with softmax.

\subsection{Patches to labels}\label{sec:patches2labels}   

Separately from the text, we \rev{trained a deep neural network model} to learn an association between \rev{just} the visual features and category and tag labels.

\textbf{Working with large images:} Since we have categories and tags for all the images in the training data, a first attempt might be to directly learn to predict the category or tag from the whole image. However, the infographics are large images often measuring beyond 1000x1000 pixels. Resizing the images reduces the resolution of visual elements which might not be perceivable at small scales. In particular, relative to the full size of the infographic, many of the pictographs take up very little real-estate but could otherwise contribute to the label prediction.
A fully convolutional approach with a batch of such large images was infeasible in terms of memory use. 
\rev{Our approach was to use a bag of sampled patches to represent the image. To sample the patches, we tried both random crops and object proposals from Alexe et al.~\cite{alexe2012measuring}.}
%As a result, we sampled the images using both random crops of a fixed size and object proposals from Alexe et al.~\cite{alexe2012measuring}. We ran the full images resized as a prediction baseline. %We also tried object proposals from 

\textbf{Multiple instance learning (MIL) prediction:} 
Given a category or tag label, we expect that specific parts of the infographic may be particularly revealing of that label, even though the whole infographic may contain many diverse visual elements. %A multiple instance learning (MIL) approach is appropriate in this case. 
In MIL, the idea is that we may have a bag of samples (in this case, image patches) to which a label corresponds. The only constraint is that at least one of the samples correspond to the label; the other samples may or may not be relevant. 

%A multiple instance learning (MIL) approach is appropriate for our task of combining the predictions of multiple samples from an image. 
We used the deep MIL formulation from Wu et al.~\cite{wu2015deep} for learning deep visual representations. We passed each sampled patch from an infographic through the same convolutional neural network architecture, and aggregated the hidden representations to predict a label for the whole bag of patches (depicted in the blue box in Fig.~\ref{fig:architecture}). For aggregating the representations, we tried both element-wise \emph{mean} and \emph{max} at the last hidden layer before the softmax transformation, but found mean worked better. As with the text model, \rev{we trained separate models for multi-class category prediction and multi-label tag prediction.}
%solved either a multi-class category prediction problem, or a multi-label tag prediction problem.

\textbf{Feature learning with patches:} We sampled 5 patches from each infographic and resized each to 224x224 pixels for input into our convolutional neural network. For feature learning, we used ResNet-50~\cite{he2016deep}, a residual neural network architecture with 50 layers, initialized by pretraining on ImageNet~\cite{russakovsky2015imagenet}. We retrained all layers of this network on 26K infographics with ground truth labels.

\begin{figure*}
\caption{Examples of how text and visual features can work together to predict the tags for an image. (a) In ``\emph{Microblogging iceberg}", visual features activate on the water and boats and predict \emph{\#travel}. The text features disambiguate the context, predicting \emph{\#social media}. Conditioned on this predicted tag, the visual features activate on the digital device icons. (b) In this comic about \emph{\#love and sex}, both textual and visual features predict \emph{\#humor}, a correct tag nevertheless. (c) In this infographic about ``\emph{Dog names}", most of the text lists dog names, specialized terms that the text model can not predict the correct tag \emph{\#animal} from. The visual features activate on the dog pictographs and make the correct tag prediction.}
\centering
\includegraphics[width=0.7\linewidth]{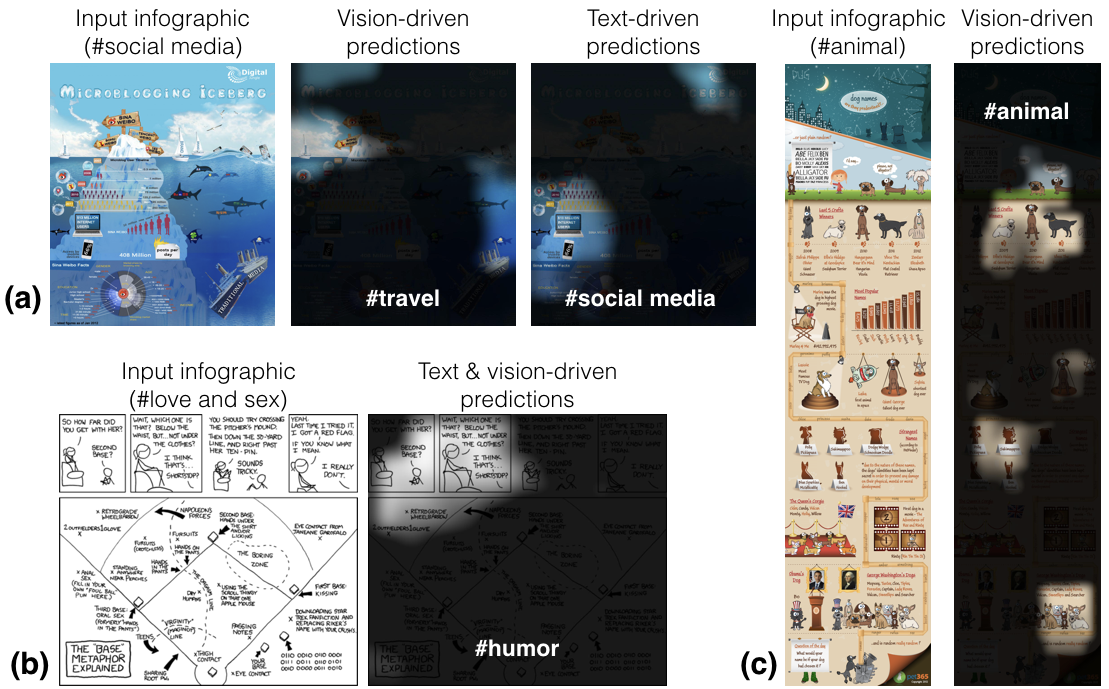}
\label{fig:vis-text-together}
\end{figure*}

\subsection{Labels to visual hashtags}\label{ssec:labels2hashtags}

The text in an infographic is often the strongest predictor of the topic matter, achieving significantly better accuracies at predicting the category and tags of infographics than the visual features alone (Sec.~\ref{sec:results}). Driven by these results, we make our initial label (category and tag) predictions using the text features. The predictions in turn constrain the visual network to produce activations for the target label. 

At inference time, we sample \rev{3500} random crops per infographic and compute the confidence, under the visual classifier, of the target label. We assign this confidence score to all the pixels within the patch, and aggregate per-pixel scores for the whole infographic. After normalizing these values by the number of sampled patches each pixel occurred in, we obtain a heatmap of activations for the target label. We use this activation map both to visualize the most highly activated regions in an infographic for a given label, and to extract visual hashtags from these regions.

For automatically extracting visual hashtags, we threshold the activation heatmap for each predicted tag, and identify connected components as proposals for regions potentially containing visual hashtags. These are cropped and passed to the SharpMask segmentation network~\cite{SharpMask}. Finally, visual hashtags corresponding to the predicted textual tags for an input infographic are obtained by cropping tight bounding boxes around SharpMask's proposals from the original images (Fig.~\ref{fig:vishashtags}).

\subsection{Technical details}

\textbf{Text model:} For category prediction, the mean word2vec representation of an infographic was fed through a 300-dimensional fully-connected linear layer, followed by a ReLu, and a 27-dimensional (including a background class) fully-connected output layer.
%Because a single 300-dimensional feature vector was used to represent each infographic, 
The feature vectors of all 29K training images fit in memory and could be trained in a single batch, with a softmax cross-entropy loss. For tag prediction, the output layer was 391-dimensional and was passed through a sigmoid layer. Given the multi-label setting, this network was trained with binary cross-entropy (BCE) loss and one-hot encoded target vectors. Both networks were trained for 20K iterations with a learning rate of $1e-3$.

\textbf{Visual model:} 
%Bags of 5 patches performed best for aggregating visual information from infographics. \rev{We report results of experiments with different patch numbers in the Supplement.}
We used bags of 5 patches for aggregating visual information from infographics.
%We also tried bags of 3 patches in batches of 33, patches of 10 in batches of 10. We used a single patch in a batch of 50 infographics as a baseline. 
\rev{We tried bags of random patches and bags of objectness proposals~\cite{alexe2012measuring}. Rather than the raw objectness proposals with varied aspect ratios, we took a tight-fitting square patch around each objectness proposal. We found this improved results.} %  (Supplement)

As in the text model, we trained category classification with a softmax cross-entropy loss with 27-dimensional target vectors, and tag prediction with a BCE loss applied to 391-dimensional sigmoid outputs.
We used a momentum of $0.9$ and weight decay of $1e-4$. Our learning rate was initialized at $1e-2$. \rev{For category prediction, we updated the learning rate every epoch, and stopped training after 5 epochs. For tag prediction, we updated the learning rate every 50 epochs for 500 epochs. Tags were more specific and also much more unbalanced than category labels, so the model needed to train for significantly longer to see enough patch samples for different tags.}
% and dropped by a factor of 10 every 10 epochs, for a total of 200 epochs. %\textcolor{blue}{TODO: rerun with a single setting of number of epochs}

\textbf{Activation maps:} To discover maximally activated image regions for a given label, \rev{3500} multi-scale crops were used. To generate each crop, we sampled a random coordinate value for the top left corner of the crop, and a side length equal to 10-40\% of the minimum image dimension.

\section{Results}\label{sec:results}

We evaluate the ability of our full system to \rev{(1)} predict category and tag labels for infographics and \rev{(2) to extract visual hashtags from images: visual regions or icons relevant to the visualization topic.} Predicting the category is a high-level prediction task about the overall topic of the infographic. Predicting the multiple tags for an infographic is a finer-grained task of discovering sub-topics. We solve both tasks, and present results of our text and visual models. 

\begin{figure}
\caption{Examples of automatic text and visual hashtag generation. (a) Text features predict half the ground truth tags correctly, and the visual model discovers associated visual regions in the infographic. Unique visual hashtags are automatically retrieved for each tag. (b) Text features predict the ground truth tags correctly, and visual features discover visual hashtags. In this specific example, there is not a one-to-one mapping between textual and visual hashtags. Similar visual areas are activated for these text tags.}
\centering
(a)\includegraphics[width=0.7\linewidth]{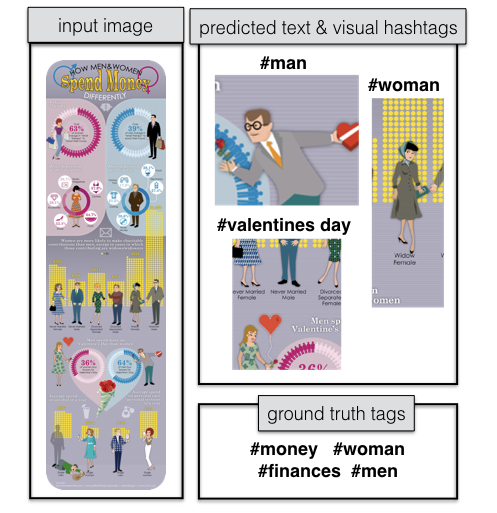}
(b)\includegraphics[width=0.7\linewidth]{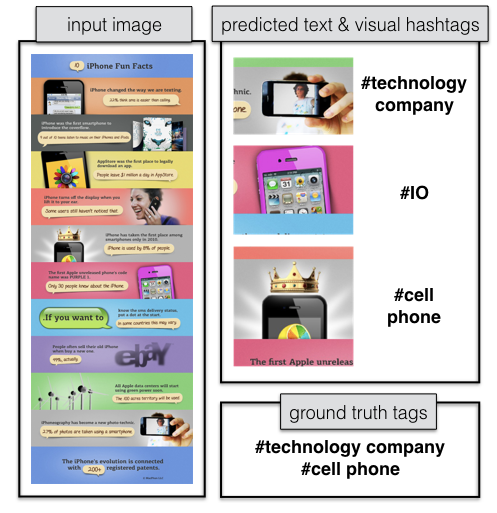}
\label{fig:visualhashtagspipeline}
\end{figure}

Given the text model's tag predictions, the visual model that learned to associate visual concepts with tags is used for finding the relevant visual areas, and to extract visual hashtag proposals (Fig.~\ref{fig:visualhashtagspipeline}). \rev{To evaluate these proposals, we collected human ground truth. For a total of 650 image-tag pairs, participants boxed image regions corresponding to the provided tag (Fig.~\ref{fig:turk}). We compare our model's visual hashtag proposals to these ground truth bounding boxes.}

\subsection{Category prediction}

\textbf{Evaluation:} For each infographic, we measured the accuracy of predicting the correct ground truth category out of 26\rev{, within the top 1, 3, and 5 most confident predictions.} 

\textbf{Quantitative results:} Chance level for our distribution of infographics across categories was 15.4\%.
We achieved \rev{46\%} top-1 accuracy at predicting the category using our text model (Table~\ref{tab:catresults}). 

The purely vision-driven predictions are provided as a comparison point, although the final label predictions are performed using the text features. The text tends to contain a lot more information, while not all concepts can be communicated visually. 
%The visual features are not intended to be comparable to the text features, as the text tends to contain a lot more information. 
The best performing visual model used a bag of random patches in a MIL framework (as in Fig.~\ref{fig:architecture}). Mean aggregation outperformed max aggregation for category prediction (\emph{Vis-rand-mean} better than \emph{Vis-rand-max}). Random crops outperformed objectness proposals (\emph{Vis-rand-mean} better than \emph{Vis-obj-mean}). \rev{We hypothesize this to be the case because each time we sampled random crops from an image, our model was exposed to new visual regions, whereas the number of objectness proposals was a limited sample of patches from an image. In other words, our model received more diverse training data in the random crops case.}
%, which we hypothesize is the case because they were more consistent, whereas object proposals had diverse aspect ratios and sizes, sometimes too small to capture meaningful visual features. 
The patch-based predictions were similar to, or better than, the full visualization resized (\emph{Vis-resized}). A patch-based approach is naturally better suited for sampling regions for visual hashtag extraction. We also tried to combine text and visual features directly during training but did not achieve gains in performance above the text model alone, indicating that it is a sufficiently rich source of information in most cases. 
%More baselines are provided in the Supplement.

%\textcolor{blue}{TODO: if time allows, (for paper:) recompute objectness with square boxes, (for supplement:) recompute text model with glove embedding}

\begin{figure}
\caption{Some of the highest activating patches per category. The visual network trained to predict category labels for whole images assigned high confidence (under the respective categories) to these randomly-sampled patches from unique infographics.}
%\centering
\includegraphics[width=0.8\linewidth]{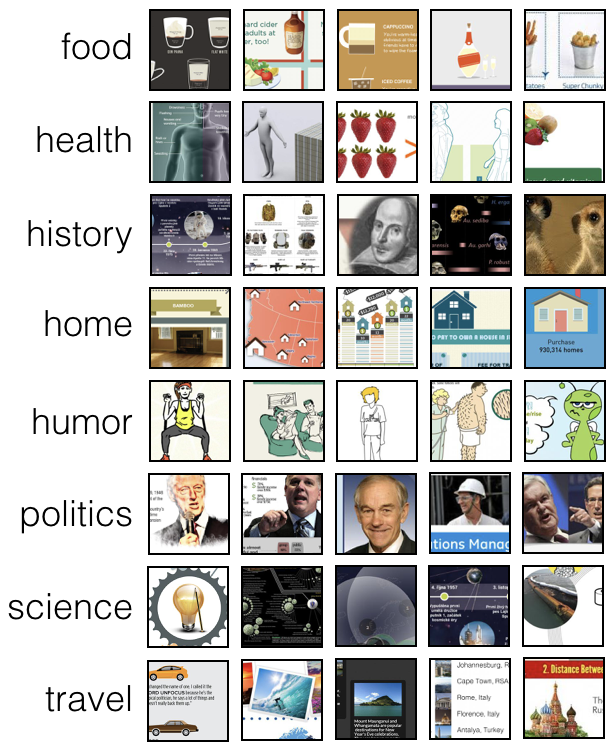}
\label{fig:topcat}
\end{figure}

\textbf{Top activations per category:}
%To validate that our visual network trained to predict categories learned meaningful features, we visualize the top 10 patches that received the highest confidence under each category. We provide the patches with the highest confidence under the classifier for each of the 26 categories in the Supplement, and a subset of patches and categories in Fig.~\ref{fig:topcat}. 
To validate that our visual network trained to predict categories learned meaningful features, we visualize the top patches that received the highest confidence under a few different categories (Fig.~\ref{fig:topcat}). 
These patches were obtained by sampling 100 random patches from each image, storing the single patch that maximally activated for each category per image, and outputting the top patches across all images.
%In Fig.~\ref{fig:topcat} we visualize 5 patches for a subset of categories that received the highest confidence under the visual network trained to recognize categories. The top 10 patches for all 26 categories are included in the supplemental. We feed 100 random patches from each image into the network and store the patch that maximally activates each category from each image...then we take the 10 highest patches for each category...note that every patch for a given category comes from a different test image.

% Patches discovered by the visual network to have highest confidence under some sample category labels

\subsection{Tag prediction}\label{sec:tagprediction}

\textbf{Evaluation:} Each infographic in our 29K dataset comes with an average of 1-9 tags. At prediction time, we generate 1, 3, and 5 tags, and measure precision and recall of these predicted tags at capturing all ground truth tags for an image, for a variable number of ground truth tags. %A single threshold might not be appropriate for all images. %\textcolor{blue}{TODO: need to measure tag accuracy at outputting all tags above threshold, for a single fixed threshold}. 
%We evaluate the tag prediction in 2 ways: (1) Average precision at predicting one of the ground truth tags. Here we predict $N=1$ tags. (2) Average precision at predicting all the ground truth tags, making as many predictions as there are ground truth tags. Here we predict $N=K$ tags, where $K$ varies per infographic between 1-9.

\textbf{Quantitative results:} We achieved \rev{48.2\%} top-1 average precision at predicting at least one of the tags for each of our infographics, since all the infographics in our dataset contain an average of 2 tags (Table~\ref{tab:tagresults}). Since tags are finer-grained than category labels, it is often the case that some word in the infographic itself maps directly to a tag. Using this insight, we add a simple automatic check: if any of the extracted words exactly match any of the 391 tags, we snap the prediction to the matching tags (\emph{Word2Vec-snap}). Without this additional step, predicting top-1 tag achieves an average prediction of \rev{30.1\%} using text features. %We provide the visual model scores for reference, although they are not directly comparable.

\rev{\textbf{Text modeling baselines:} We computed several other representations of the extracted text (Table \ref{tab:tag-prediction-from-text}).
% compared the mean word2vec representation to several other
We used a voting scheme (\emph{Word2Vec-voting}) by voting for the closest text tag, in word2vec embedding space, for each word in the extracted text, and predicting the top-voted tags.
%To compute \emph{Word2vec-voting} we identified the closest text tag, in word2vec embedding space, for each word in the extracted text, and cast a vote for that tag. The top voted text tags were predicted. 
We also computed the \emph{Tweet2Vec} \cite{dhingra-EtAl:2016:P16-2} representation of the extracted text, as well as the mean of the Glove representations~\cite{pennington2014glove} of all the words (\emph{Glove-mean}).
%Secondly, we tried using the mean glove representation of the words as the text features. Finally, we used the tweet2vec embedding as described in \cite{dhingra-EtAl:2016:P16-2} as the text feature for the extracted text. As shown in the results, 
Using the mean word2vec as the text features (\emph{Word2Vec-mean}) gave the best results for tag prediction.}

\textbf{Text can disambiguate visual predictions:} In some infographics, visual cues for particular tags or topics may be missing (e.g., for abstract concepts), they may be misleading (as visual metaphors), or they may be too numerous (in which case the most representative must be chosen). In these cases, label predictions driven by text are key, as in Fig.~\ref{fig:vis-text-together}a, where visual features might seem to indicate that the infographic is about icebergs, or ocean, or travel; in this case, however, iceberg is used as a metaphor to discuss microblogging and social media. Our text model is able to pick up on this, and direct the visual features to activate in the relevant regions.

\subsection{Visual hashtag proposals}\label{sec:proposals}

\begin{figure}
\caption{\rev{We collected ground truth visual hashtags by asking dozens of participants to box any visual regions corresponding to a provided tag. For evaluating our automatic model, ground truth was collected for a total of 650 image-tag pairings.}}
%Two tasks used to gather ground-truth human annotations for (a) text hashtags, and (b) visual hashtags. In the first task (above), participants could scroll over an image to zoom in and inspect parts of it in order to generate a set of representative hashtags for the image. In the second task (below), participants annotated visual regions matching text tags.}
\centering
%(a)\includegraphics[width=0.8\linewidth]{Figures/mturk1.png}
%(b)
\includegraphics[width=1\linewidth]{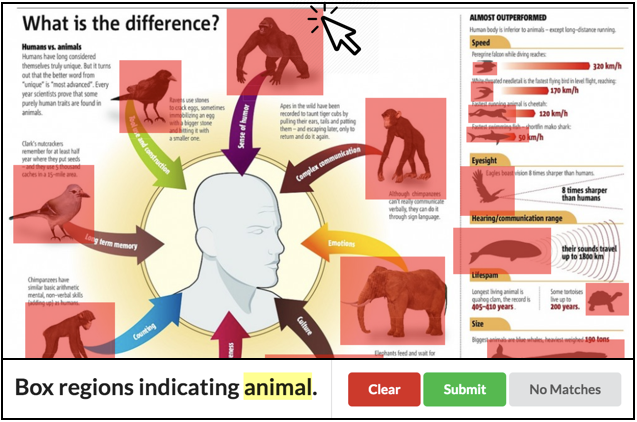}
\label{fig:turk}
\end{figure}

\textbf{Collecting ground truth:} The \emph{Visually} data comes with image-level categories and tags. Because a goal of this paper is to discover visual hashtags - individual elements \emph{within} infographics that correspond to the different labels - we wanted to measure how humans complete this task. We designed an interface in which participants are given an infographic and a text tag, and are asked to mark bounding boxes around all non text-regions (e.g., pictographs) that contain a depiction of the tag (Fig.~\ref{fig:turk}). %We used the designer-assigned tags from the \emph{Visually} dataset as the target tags. 
If an image had multiple tags, it would be shown multiple times but to different users, with unique image-tag pairings. %Because this was a more difficult task requiring a good understanding of the instructions, we opted for data from undergraduate students rather than MTurk workers. 
We collected a total of 3655 bounding boxes (ground truth visual hashtags) for the 330 images from 43 undergraduate students. Each image was seen by an average of 3 participants and we obtained an average of 4 boxes per image.

\textbf{Evaluation:} On average, infographics had 2 ground truth tags, with a total of 650 unique image-tag pairs for which participants annotated visual hashtags. Of these 650 pairings, participants indicated that 119 (18\%) did not have corresponding visual features. In these cases, the hashtag had no visual counterpart and could perhaps only be inferred from the text of the infographic. %, as we found before. %Visual features alone can not predict all the tags and categories of visualizations. %, which explains why text achieves significantly higher performances at tag prediction. reinforcing our previous findings

\begin{figure*}
\centering
\caption{Some sample visual hashtag extraction results. We show multiple steps of our pipeline: given a text tag, the activation heatmap indicates the image regions that our visual model predicts as most relevant. This heatmap is then passed to our pipeline that extracts visual hashtags, using objectness and text detection to filter results. The final extracted hashtags are included. We overlay our proposed visual hashtags (in blue) with human-annotated bounding boxes (in red) of relevant visual regions to the text tag.}
\includegraphics[width=0.5\linewidth]{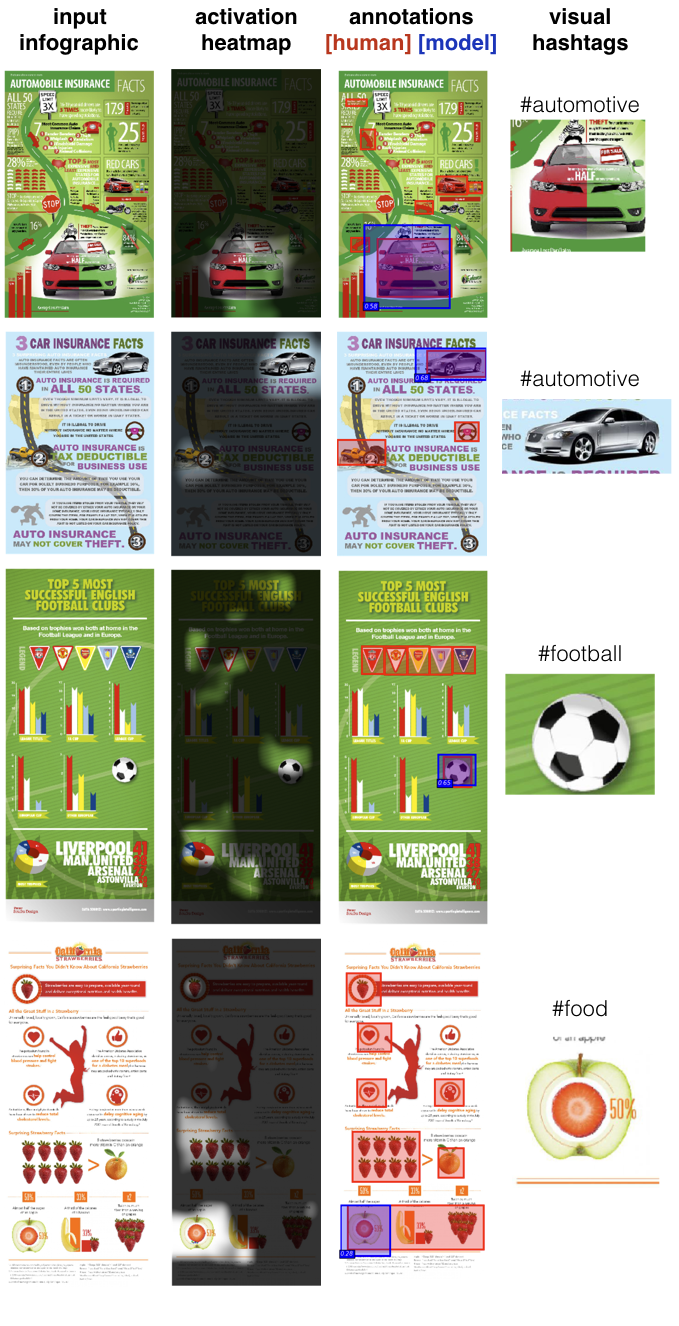}
\includegraphics[width=0.435\linewidth]{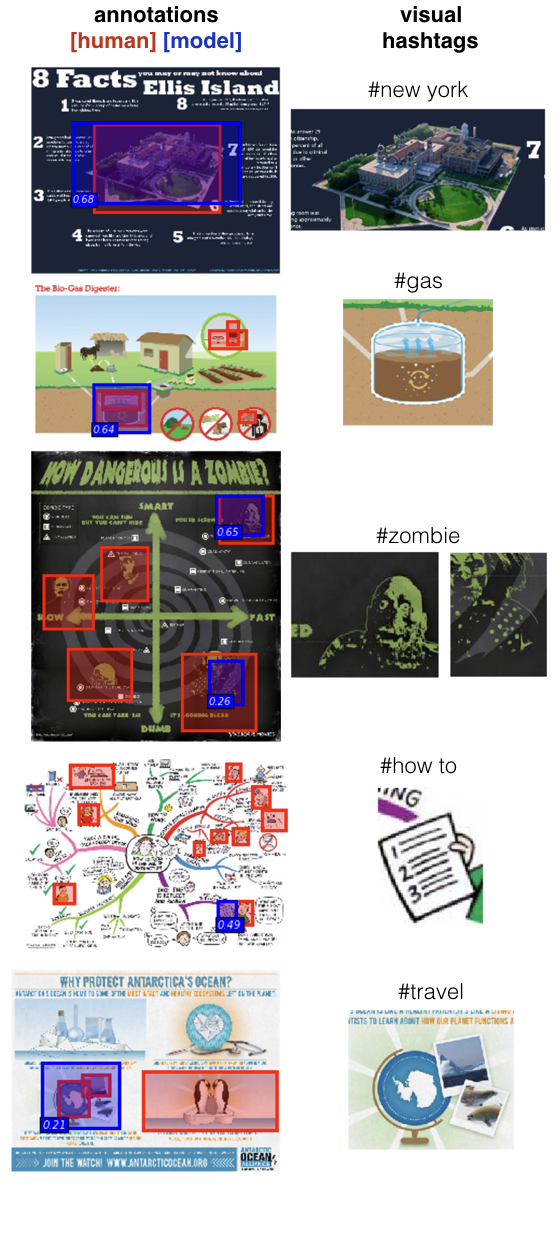}
\label{fig:hashtag_examples}
\end{figure*}

We evaluated the remaining 531 image-tag pairs with participant annotations (ground truth hashtags). \rev{We fed each of these image-tag pairs to our pipeline to obtain predicted visual hashtags (Sec.~\ref{ssec:labels2hashtags}) and computed the intersection-over-union (IOU) of each of our predicted hashtags with participant annotations. We report only the single highest-confidence prediction for each image-tag pair (Table~\ref{tab:vishashtag}). The confidence of our proposals is measured as the mean activation value of our visual model within the hashtag bounding box. See Fig.~\ref{fig:hashtag_examples} for examples of our predicted hashtags overlaid with participant annotations.

Our pipeline was constructed for high-precision as opposed to high-recall, because our goal is to produce a reasonable visual hashtag for an image-tag pair, rather than all possible hashtags. Therefore, our evaluation measures the percent of predictions that overlap with at least one of the human annotations. We report precision as the percent of predicted hashtags that have an $IOU > 0.5$ with at least one ground truth hashtag (in an image-tag pair). This threshold was chosen because it is most commonly used in the object detection literature~\cite{everingham2010pascal}. 

To contrast with precision, we also report the total percent of image-tag pairs for which a successful proposal with $IOU > 0.5$ was generated (\emph{Acc.}). In this case, for any image-tag pair for which a proposal was not generated, IOU is set to be 0.
%We also report the average IOU across all proposals (Table~\ref{tab:vishashtag}). 
%We achieved a precision of 12.7\%.

\textbf{Object proposal baselines:} Our average precision of 15.2\% and accuracy of 9.4\% beat other approaches on the task of outputting a visual hashtag proposal for a given image-tag pair (Table~\ref{tab:vishashtag}). We took the highest-confidence object proposals from Alexe et al.~\cite{alexe2012measuring} (\emph{Objectness}) and Pinheiro et al.~\cite{SharpMask} (\emph{SharpMask}). We also used a top-performing neural network model of saliency~\cite{pan2016shallow} (\emph{SalNet}) in place of our visual model's activation map, and ran it through the same post-processing pipeline as outlined in Sec.~\ref{ssec:labels2hashtags} to obtain visual hashtag proposals. The benefits our activation map has over saliency is that saliency is tag-agnostic and will always output the same map for an image. Our visual model is conditioned on a particular tag label and activates in regions of a design that are most predictive of the label. For a comparison to another weakly-supervised approach, we adjust our network to have an average pooling layer at the end, as in CAM~\cite{zhou2016learning}.
As a chance baseline we report the performance of random crops (\emph{Random}). 

\textbf{Increasing accuracy:} When we take into account all image-tag pairs, the average percent of instances for which the predicted hashtag overlaps the ground truth with an $IOU > 0.5$ drops to 9.4\% (from a precision of 15.2\%). Our approach fails to output proposals for 38\% of the image-tag pairs. Most of the filtering happens at the SharpMask stage, where region proposals from the visual activation map are passed to SharpMask for refinement. If SharpMask does not find an object candidate in an image region, that region is discarded. As a stand-alone method, \emph{SharpMask} fails to output proposals for 34\% of image-tag pairs. SharpMask is also used as a post-processing step for the \emph{SalNet} model. In comparison, \emph{Objectness} generates a candidate for all images.
%The selectivity of SharpMask can also be seen in Table~\ref{tab:vishashtag} where SharpMask appears as a stand-alone method. 
We can increase the percent of proposals returned by adding a fallback option to our method (\emph{Ours-fallback}): even if SharpMask discards all candidates, return the most confident candidate. This allows us to guarantee proposals for all images, at the cost of lower precision.
}

\begin{comment}
\begin{table}[]
\small
\caption{}
\centering
\begin{tabu} to \linewidth {c c c c}
\toprule
Model 			& Prec. 	& Avg. IOU	& Success rate		\\ \midrule
Ours			& 15.2\%	& 0.18		& 62\%		\\ 
Ours-fallback	& 10.5\%    & 0.14      & 100\%		\\
SalNet			& 10.9\%    & 0.14		& 63\%      \\ 
Objectness		& 9.0\%		& 0.13		& 100\%		\\ 
SharpMask		& 8.6\% 	& 0.10		& 66\%		\\ 
Random			& 0.9\%		& 0.06		& 100\%		\\ 
\bottomrule
\end{tabu}
\label{tab:vishashtag}
\end{table}
\end{comment}

\begin{table}[]
\small
\caption{\rev{We measure the precision of different strategies for proposing visual hashtags across different image-tag pairings.}}
\centering
\begin{tabu} to \linewidth {c c c}
\toprule
Model 			& Prec. 	& Acc.		\\ \midrule
Ours			& 15.2\%	& 9.4\%		\\ 
Ours-fallback	& 10.5\%    & 10.5\%	\\
SalNet~\cite{pan2016shallow}	& 10.9\%    & 7.0\%     \\ 
Objectness~\cite{alexe2012measuring}		& 9.0\%		& 9.0\%		\\ 
SharpMask~\cite{SharpMask}		& 8.6\% 	& 5.6\%		\\ 
CAM~\cite{zhou2016learning}		& 5.4\%		& 2.8\%		\\
Random			& 0.9\%		& 0.01\%	\\ 
\bottomrule
\end{tabu}
\label{tab:vishashtag}
\end{table}

\begin{comment}
\rev{TODO: replace the following with IOU analysis.} For each image-tag pair, we masked out all the image regions annotated by participants for a given image-tag pair, producing a binary mask. We then measured the mean activation value in this binary mask predicted by our visual model. Specifically, we normalized the visual activation heatmap to be zero-mean and computed the average normalized heatmap value in the region annotated by participants. For 65\% of the tags, our visual model activated above chance in the human-annotated region. We examined which tags our visual model best localized based on the mean activation value of those visual regions across images. The best localized tags include \emph{home, automotive, food, weather, new york}, and among the worst localized tags are more abstract tags such as \emph{savings, wellness, investment, happiness, medium, stress}. More details are in the Supplement.
\end{comment}

\begin{comment}
- >= 43 participants (some people left email field blank)
- 3655 total boxes
- 634 total tag/image pairs considered
- 327 filenames considered
- 3655/43 = ~85 bounding boxes drawn per participant
- 634/43 = ~14.74 images per participant
\end{comment}

%\subsection{Analysis}

%\textcolor{blue}{Points to make: external validation; user study; similar level of abstraction; tags are reproducible; people asked to generate hashtags reproduce our ground truth tags }

%\section{Discussion}\label{sec:discussion}
%\textbf{Text can disambiguate visual predictions:}  
%\textcolor{blue}{TODO: include figure where visual features alone can not predict label, text helps.}
%\textcolor{blue}{TODO: describe again the power of in-context text to supervise visual features.}

\begin{comment}
\begin{table}[h]
\small
\centering
\caption{Results on category prediction \textcolor{blue}{TODO: (1) compute top-3 for visual nets; (2) compute chance level; (3)}} \label{tab:catresults}
\begin{tabular}{| c | c | c | c |}
\hline
  Model & Top-1 & Top-3 & Top-5 \\ \hline
  Text-mean & 43.2\% & 69.3\% & 79.4\% \\ \hline 
  Vis-resized & 28\% & & 64\% \\ \hline
  Vis-rand-mean & 30.3\% & & 63.8\% \\ \hline
  Vis-rand-max & 27.4\% & & 63.1\% \\ \hline
  Vis-obj-mean & 27.5\% & & 61.2\% \\ \hline
  Vis-obj-max & 27.3\% & & 60.1\% \\ \hline
  Chance & && \\ \hline
\end{tabular} 
\end{table}
\end{comment}

% NOTE: Zoya updated these table numbers on July 12
\begin{table}[h]
\small
\centering
\caption{Results on category prediction using visual features only. Models sorted in order of Top-1 performance.} \label{tab:catresults}
\begin{tabu} to \linewidth {c c c c c}
\toprule
  Model & Top-1 & Top-3 & Top-5 \\ \midrule
  Vis-rand-mean & 26.5\% & 49.8\% & 62.3\% \\ 
  Vis-rand-max & 25.6\% & 48.8\% & 62.2\% \\ 
  Vis-obj-mean & 25.4\% & 47.8\% & 60.8\% \\ 
  Vis-obj-max & 24.2\% & 47.8\% & 61.4\% \\ 
  Vis-resized & 23.7\% & 47.2\% & 60.0\% \\ 
  Chance &15.4\% & 33.6\% & 47.5\%\\ \bottomrule
\end{tabu} 
\end{table}

\definecolor{LightCyan}{rgb}{0.88,1,1}
\definecolor{LightGray}{gray}{0.95}
\definecolor{DarkGray}{gray}{0.2}

\begin{table}[h]
\small
\centering
\caption{Results on category prediction using extracted text only. Models sorted in order of Top-1 performance.} \label{tab:catresults}
\begin{tabu} to \linewidth {c c c c}
\toprule
   Model & Top-1 & Top-3 & Top-5 \\ \midrule
  %\rowcolor{LightGray}
  Word2Vec-mean & 46\% & 72.3\% & 83.2\% \\ %  top-k 23.2\%
  Glove-mean & 23.9\% & 45.4\% & 58.5\%  \\ %\midrule
  %\rowcolor{LightGray}
  Tweet2Vec & 24.4\% & 45.9\% & 78.8\%  \\ %\midrule
  Word2Vec-voting  & 10\% & 39.9\% & 52.94\% \\  %\\ %\midrule
\bottomrule 
\end{tabu} 
\end{table}

%\rowcolor{LightGray}
%  \textcolor{DarkGray}{\emph{Recall}}

\begin{table}[h]
\small
\centering
\caption{Results on textual tag prediction using visual features only. Models sorted in order of Top-1 performance.} \label{tab:tagresults}
\begin{tabu} to \linewidth {c c c c c}
\toprule
  Model & Acc. & Top-1 & Top-3 & Top-5 \\ \midrule
  \rowcolor{LightGray}
  Vis-rand-max & prec & \rev{15.1\%} & \rev{9.8\%} & \rev{7.9\%} \\ 
  \rowcolor{LightGray}
  & rec & \rev{8.2\%} & \rev{15.1\%} & \rev{20.2\%}\\ 
    Vis-rand-mean & prec & \rev{14.3\%} & \rev{9.9\%} & \rev{7.7\%} \\ 
  & rec & \rev{7.7\%} & \rev{15.2\%} & \rev{19.6\%} \\ 
%  Vis-obj-mean & prec & \rev{11.0\%} & \rev{8.1\%} & \rev{6.6\%}
% \\ 
%  & rec & \rev{6.2\%} & \rev{12.6\%} & \rev{16.9\%} \\ \hline
%  Vis-obj-max & prec & \rev{10.5\%} & \rev{7.6\%} & \rev{6.3\%}
% \\
%  & rec & \rev{6.0\%} & \rev{11.7\%} & \rev{16.2\%} \\ \hline
\rowcolor{LightGray}
  Vis-obj-max & prec & \rev{13.4\%} & \rev{9.1\%} & \rev{7.3\%}
 \\
 \rowcolor{LightGray}
  & rec & \rev{7.5\%} & \rev{13.9\%} & \rev{18.5\%} \\ 
  Vis-obj-mean & prec & \rev{13.0\%} & \rev{9.2\%} & \rev{7.4\%}
 \\ 
  & rec & \rev{7.1\%} & \rev{14.1\%} & \rev{19.0\%} \\ 
  \rowcolor{LightGray}
  Vis-resized & prec & 12.1\% & 8.2\% & 6.8\% \\ 
  \rowcolor{LightGray}
  & rec & 6.5\% & 13.1\% & 17.8\% \\ 
  Chance & prec & 8.7\% & 6.4\% & 5.5\% \\ 
  & rec & 5.1\% & 10.3\% & 14.3\%\\ 
  \bottomrule
\end{tabu} 
\end{table}

\begin{comment}
\begin{tabular}{| c | c | c | c | c |}
\hline
  Model & Acc. & Top-1 & Top-3 & Top-5 \\ \hline
  Vis-rand-mean & prec & 12.2\% & 8.4\% & 6.9\% \\ 
  & rec & 6.7\% & 13.1\% & 17.8\% \\ \hline
  Vis-rand-max & prec & 12.2\% & 8.4\% & 6.5\% \\ 
  & rec & 6.8\% & 13.0\% & 16.8\% \\ \hline
  Vis-resized & prec & 12.1\% & 8.2\% & 6.8\% \\ 
  & rec & 6.5\% & 13.1\% & 17.8\% \\ \hline
  Vis-obj-mean & prec & 11.4\% & 8.1\% & 6.6\% \\ 
  & rec & 6.4\% & 12.6\% & 17.0\% \\ \hline
  Vis-obj-max & prec & 11.1\% & 8.1\% & 6.4\% \\
  & rec & 6.1\% & 12.5\% & 16.4\% \\ \hline
  Chance & prec & 8.7\% & 6.4\% & 5.5\% \\ 
  & rec & 5.1\% & 10.3\% & 14.3\%\\ \hline
\end{tabular} 
\end{table}
\end{comment}

\begin{table}[]
\small
\caption{Results on textual tag prediction using extracted text only. Models sorted in order of Top-1 performance.} \label{tab:tag-prediction-from-text}
\centering
\begin{tabu} to \linewidth {c c c c c}
\toprule
  Model & Acc. & Top-1 & Top-3 & Top-5 \\ \midrule
  \rowcolor{LightGray}
  Word2Vec-snap & prec & 48.2\% & 28.1\% & 20\% \\ %  top-k 23.2\% 
  \rowcolor{LightGray}
  & rec & 45.3\% & 52.6\% & 57.2\% \\ %\midrule
  Word2Vec-mean & prec & 30.1\% & 20.1\% & 15.3\%  \\ 
  & rec & 17.0\% & 32.5\% & 40.2\%  \\ %\midrule
  \rowcolor{LightGray}
  Glove-mean & prec & 29.1\% & 18.6\% & 14.2\%  \\ 
  \rowcolor{LightGray}
  & rec & 16.8\% & 29.7\% & 37.4\%  \\ %\midrule
  Tweet2Vec & prec & 6.5\% & 18.1\% & 14.0\%  \\ 
  & rec & 16.9\% & 29.5\% & 37.2\%  \\ %\midrule
  \rowcolor{LightGray}
  Word2Vec-voting & prec & 1.2\% & 7.8\% & 8.4\%  \\ 
  \rowcolor{LightGray}
  & rec & 0.6\% & 12.2\% & 21.9\%  \\ 
\bottomrule
\end{tabu}
\end{table}

\section{Conclusion}

To this point, the space of complex visual information beyond natural images has received limited attention in computer vision in the domain of classification and detection (notable exceptions include: \cite{kembhavi2016eccv,zitnick2013bringing}). We present a novel direction based on a dataset of infographics, containing rich visual media, with a mix of visual and textual features. In this paper, we showed how textual and visual elements can be used to jointly reason about the high-level topics (categories) of infographics, as well as the finer-grained sub-topics (tags). We demonstrated the power of text features in disambiguating and providing context for visual features. We presented a system whereby aside from predicting text labels, we can automatically extract iconic representative elements,  what we call ``visual hashtags". \rev{Despite never being trained to explicitly recognize objects in images, our model is able to localize a subset of the ground truth (human-annotated) visual hashtags.}

Infographics are specifically designed with a human viewer in mind, characterized by higher-level semantics, such as a story or a message. Beyond simply detecting the objects contained within them, an understanding of these infographics would involve the parsing and understanding of the included text, the layout and spatial relationships between the elements, and the intent of the designer. Human designers are experts at piecing together elements that are cognitively salient (or memorable) and maximize the utility of information. This new space of multi-media data gives computer vision researchers the opportunity to model and understand the higher-level properties of textual and visual elements of the story being told. % Visual hashtags whose meaning goes beyond object recognition, condense an entire infographic into its semantic essence. 

% This new space of multi-media data can provide computer vision researchers with the opportunity to ask new questions: what are the most semantically or important elements, what higher-level messages do they convey, what inferences will a human viewer extract, which memories will be activated.

%This space opens up many additional questions for computer vision researchers. 

{\small
\bibliographystyle{ieee}
\bibliography{egbib}
}

\begin{comment}
\clearpage
\newpage
\section{Supplemental [temporary]}

\begin{figure*}
\caption{}
%\centering
%\includegraphics[width=0.9\linewidth]{Figures/categ_examples001.png}\vspace{-2mm}
%\includegraphics[width=0.9\linewidth]{Figures/categ_examples002.png}
\includegraphics[width=0.9\linewidth]{Figures/category_examples_fullpage1.png}
\label{fig:topcat1}
\end{figure*}

\begin{figure*}
\caption{}
%\centering
%\includegraphics[width=0.9\linewidth]{Figures/categ_examples003.png}\vspace{-2mm}
%\includegraphics[width=0.9\linewidth]{Figures/categ_examples004.png}
\includegraphics[width=0.9\linewidth]{Figures/category_examples_fullpage2.png}
\label{fig:topcat2}
\end{figure*}

SAMI COMMENTS
- sort models in order of performance actually
- make sure capitalization in titles is consistent
- remove colon after “Tag prediction”
- fix hashtag / emph / quotation marks
- text vs. textual
- “. or .” ? also same for commas
- IOC in supplement
- more on paper copy

\end{comment}

\end{document}